\documentclass[runningheads]{llncs}

\usepackage{graphicx}
\usepackage{subfigure}
\usepackage{xcolor}
\usepackage{amssymb}
\usepackage{color,soul}
\usepackage{marvosym}

\begin{document}

\title{CCDWT-GAN: generative adversarial networks \\ based on color channel using discrete wavelet \\ transform for document image binarization}
\titlerunning{CCDWT-GAN}

\author{Rui-Yang Ju \inst{1,2} \textsuperscript{(\Letter)} \and
Yu-Shian Lin \inst{1}\and
Jen-Shiun Chiang \inst{1} \textsuperscript{(\Letter)} \and
Chih-Chia Chen \inst{1} \and
Wei-Han Chen \inst{1} \and
Chun-Tse Chien \inst{1}}

\authorrunning{Ju et al.}

\institute{Tamkang University, New Taipei City, 251301, Taiwan \and National Taiwan University, Taipei City, 106319, Taiwan \\
\email{\{jryjry1094791442,jsken.chiang\}@gmail.com}}

\maketitle
\vspace{-0.5cm}
\begin{abstract}
To efficiently extract textual information from color degraded document images is a significant research area. The prolonged imperfect preservation of ancient documents has led to various types of degradation, such as page staining, paper yellowing, and ink bleeding. These types of degradation badly impact the image processing for features extraction. This paper introduces a novelty method employing generative adversarial networks based on color channel using discrete wavelet transform (CCDWT-GAN). The proposed method involves three stages: image preprocessing, image enhancement, and image binarization. In the initial step, we apply discrete wavelet transform (DWT) to retain the low-low (LL) subband image, thereby enhancing image quality. Subsequently, we divide the original input image into four single-channel colors (red, green, blue, and gray) to separately train adversarial networks. For the extraction of global and local features, we utilize the output image from the image enhancement stage and the entire input image to train adversarial networks independently, and then combine these two results as the final output. To validate the positive impact of the image enhancement and binarization stages on model performance, we conduct an ablation study. This work compares the performance of the proposed method with other state-of-the-art (SOTA) methods on DIBCO and H-DIBCO ((Handwritten) Document Image Binarization Competition) datasets. The experimental results demonstrate that CCDWT-GAN achieves a top two performance on multiple benchmark datasets. Notably, on DIBCO 2013 and 2016 dataset, our method achieves F-measure (FM) values of 95.24 and 91.46, respectively.

\end{abstract}

\section{Intorduction}
\vspace{-0.2cm}
Document image binarization is a significant research topic in Computer Vision (CV). Although the traditional image binarization methods are capable of extracting textual information from regular document images, they often struggle to process degraded ancient document images, including text degradation and bleed-through \cite{kligler2018document,sulaiman2019degraded}.

In recent years, image binarization methods based on deep learning have shown remarkable performance in addressing the problems that traditional image binarization methods \cite{niblack1985introduction,otsu1979threshold,sauvola2000adaptive} cannot solve. Several methods have been proposed and achieved state-of-the-art (SOTA) performance in degraded document image binarization, such as the conditional generative adversarial network-based method \cite{zhao2019document}, the hierarchical deep supervised network \cite{vo2018binarization}, and the iterative supervised network \cite{he2019deepotsu}, which all outperform traditional image binarization methods and other deep learning-based methods \cite{guo2019nonlinear,tensmeyer2017document,yang2023gdb}.

The aforementioned image binarization methods generally have superior results when applied to grayscale documents, particularly for restoring contaminated black and white scanned ancient documents. Considering that some scanned images of ancient documents are in color, we propose generative adversarial networks based on color channel using discrete wavelet transform (CCDWT-GAN), which utilize the discrete wavelet transform (DWT) on RGB (red, green, blue) split images to binarize the color degraded documents.

This paper makes the following contributions:
\vspace{-0.1cm}
\begin{itemize}
    \item [1)] Demonstrating that applying DWT on RGB split images can improve the efficiency of the generator and the discriminator.
    \item [2)] Presenting a novel method for document image binarization that achieves SOTA performance on multiple benchmark datasets.
\end{itemize}
\vspace{-0.1cm}

The rest of this paper is organized as follows: Section \ref{related} introduces the related work of document image binarization and GANs. Section \ref{method} provides detailed information about the proposed method. Section \ref{experiments} presents a quantitative comparison with SOTA methods on benchmark datasets. Finally, Section \ref{conclusion} concludes this paper.

\vspace{-0.2cm}
\section{Related Work}
\vspace{-0.2cm}
\label{related}
There are two primary categories of document image binarization methods: traditional image binarization methods and deep-learning-based semantic segmentation methods. The traditional image binarization method involves binarizing the image by calculating a pixel-level local threshold \cite{howe2013document,jia2018degraded}. On the other hand, the deep learning-based semantic segmentation method utilizes U-Net 
\cite{ronneberger2015u} to capture contextual and location information. This method utilizes an encoder-decoder structure to transform the input image into the binarized representation \cite{he2019deepotsu,jemni2022enhance,tensmeyer2017document,vo2018binarization}.

Recently, generative adversarial networks (GANs) \cite{goodfellow2020generative} have shown impressive success in generating realistic images. Zhao \emph{et al.} \cite{zhao2019document} introduced a cascaded generator structure based on Pix2Pix GAN \cite{isola2017image} for image binarization. This architecture effectively addresses the challenge of combining multi-scale information. Bhunia \emph{et al.} \cite{bhunia2019improving} conducted texture enhancement on datasets and utilized conditional generative adversarial networks (cGAN) for image binarization. Suh \emph{et al.} \cite{suh2022two} employed Patch GAN \cite{isola2017image} to propose a two-stage generative adversarial networks for image binarization. De \emph{et al.} \cite{de2020document} developed a dual-discriminator framework that fuses local and global information. These methods all achieve the SOTA performance for document image binarization.

\begin{figure}[ht]
  \centering
  \includegraphics[width=\linewidth]{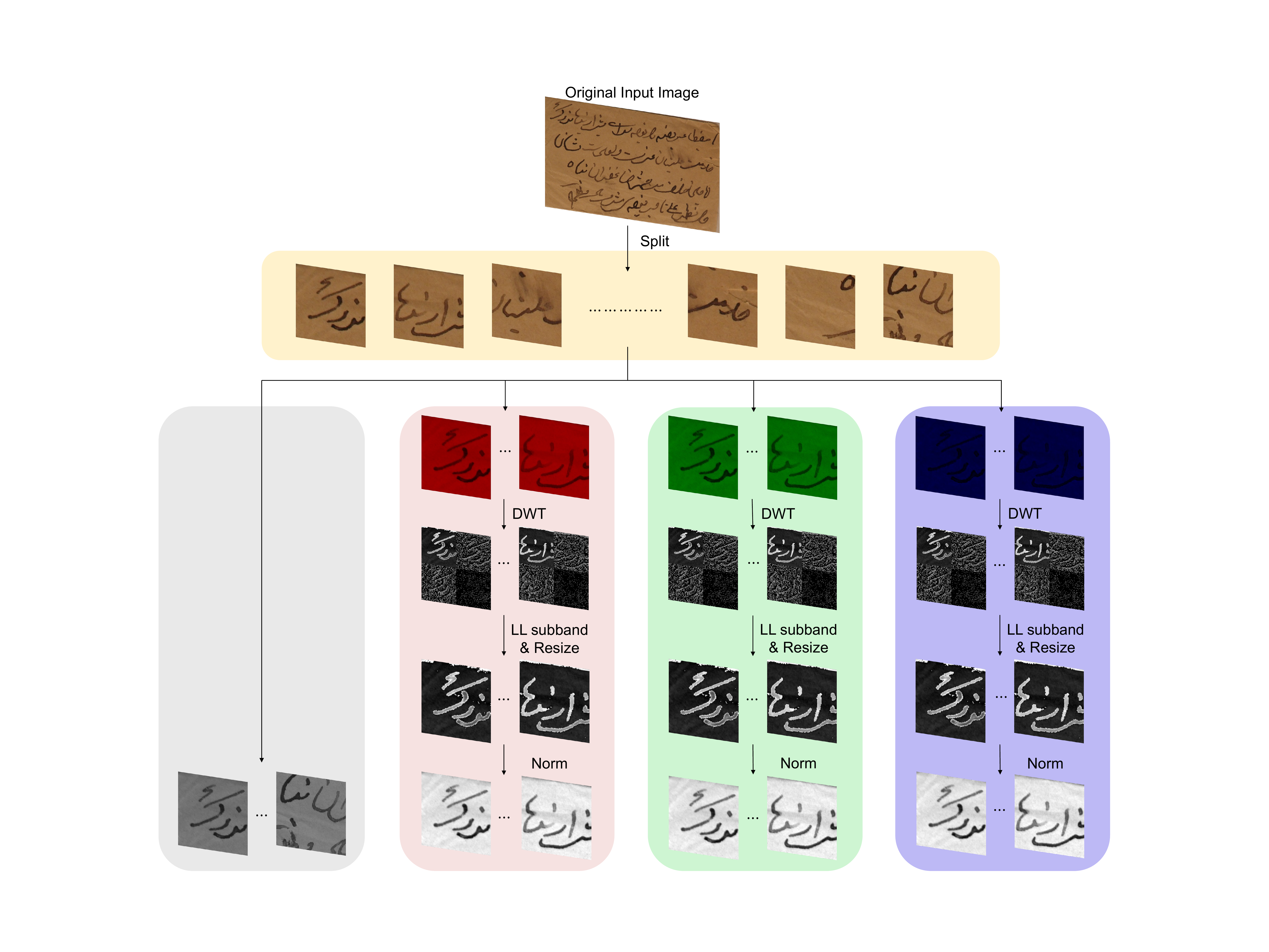}
  \caption{The structure of the proposed model for image preprocessing. The original input image is split into multiple $224\times224$ patches. After applying DWT, the LL subband images are retained from the RGB channels split images. These images are subsequently resized to $224\times224$ pixels and perform normalization.}
  \label{figure_stage_1}
  \vspace{-0.5cm}
\end{figure}

\section{Proposed Method}
\vspace{-0.2cm}
\label{method}
This work aims to perform image binarization on color degraded document images. Due to the diverse and complex nature of document degradation, our method employs CCDWT-GAN on both RGB split images and a grayscale image. The proposed method consists of three stages: image preprocessing, image enhancement, and image binarization.

\subsection{Image Preprocessing}
\vspace{-0.2cm}
In the first step, the proposed method employs four independent generators to extract the foreground color information and eliminate the background color from the image. To obtain different input images for four independent generators, we first split the RGB three-channel input image into three separate single-channel images and a grayscale image, as shown in Fig. \ref{figure_stage_1}. To preserve more information in RGB channels split images, this work applies DWT to each single-channel images to retain the LL subband images, then resizes to 224$\times$224 pixels, and finally performs normalization. There are many options to process the input image of the generator and the discriminator, such as whether to perform normalization. In section \ref{experiment}, we conduct comparative experiments to find the best option.

\begin{figure}[ht]
  \centering
  \includegraphics[width=\linewidth]{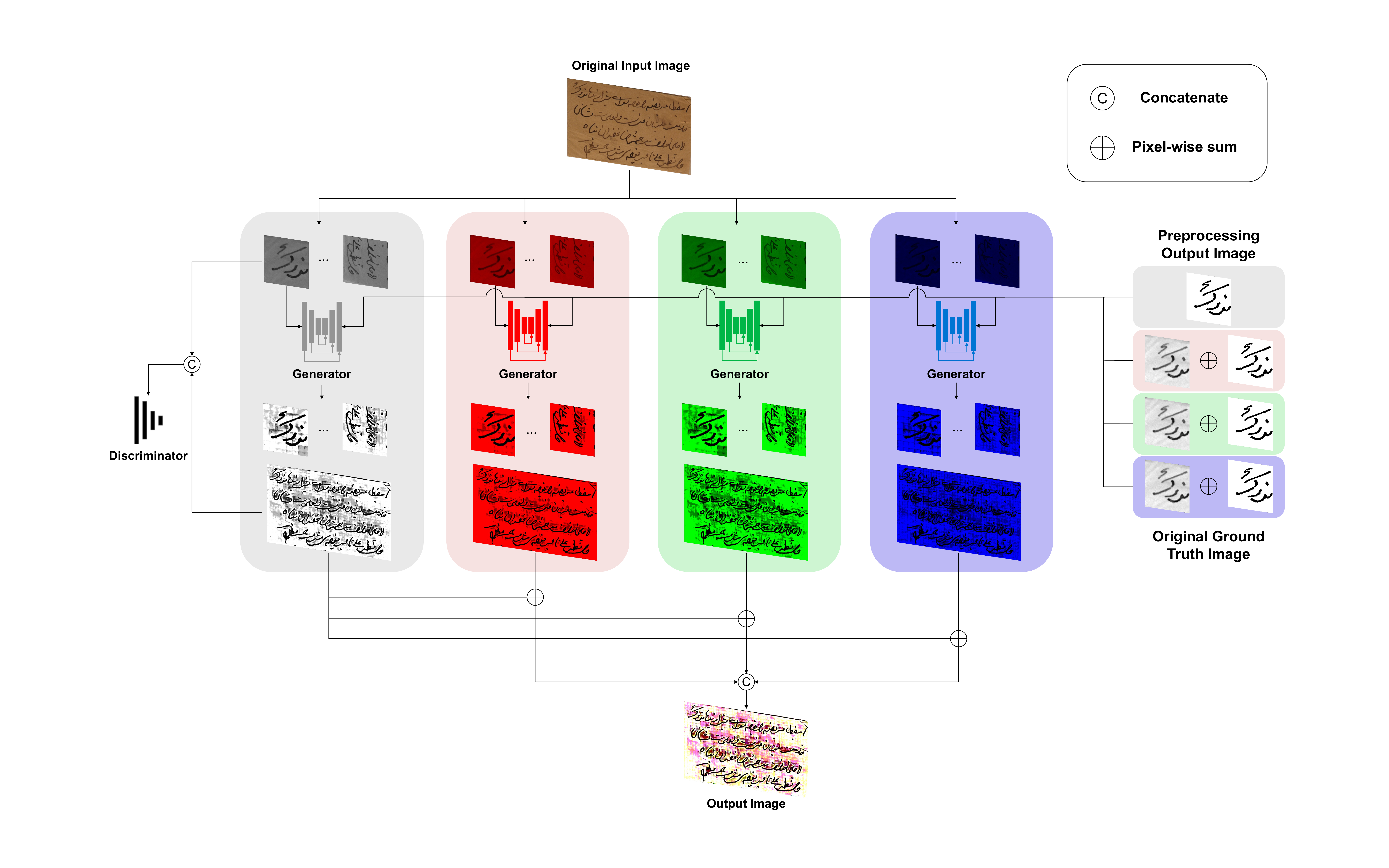}
  \caption{The structure of the proposed model for image enhancement. The preprocessing output images and the original ground truth images are summed (pixel-wise) as the ground truth images of the generator.}
  \label{figure_stage_2}
  \vspace{-0.5cm}
\end{figure}

\subsection{Image Enhancement}
\vspace{-0.2cm}
In this stage, depicted in Fig. \ref{figure_stage_2}, the RGB input image with three channels is split into three separate single-channel images and a grayscale image. Each of these image utilizes an independent generator and shares the same discriminator to distinguish between the generated image and its corresponding ground truth image. The trained network is capable of eliminating background information from the local image patches and extracting color foreground information. To extract features, we employ U-Net++ \cite{zhou2019unet++} with EfficientNet \cite{tan2019efficientnet} as the generator.

Due to the unpredictable degree of document degradation, four independent adversarial networks are used to extract text information from various color backgrounds, minimizing the interference caused by color during document image binarization. Since images with different channel numbers cannot be directly put into the same discriminator, the input of the discriminator requires a three-channel image, and the ground truth image is a grayscale (single-channel) image. As shown in the right of Fig. \ref{figure_stage_2}, the original ground truth image and the output image obtained from image preprocessing are summed at the pixel level to serve as the corresponding ground truth images.
\vspace{-0.2cm}

\begin{figure}[ht]
  \centering
  \includegraphics[width=\linewidth]{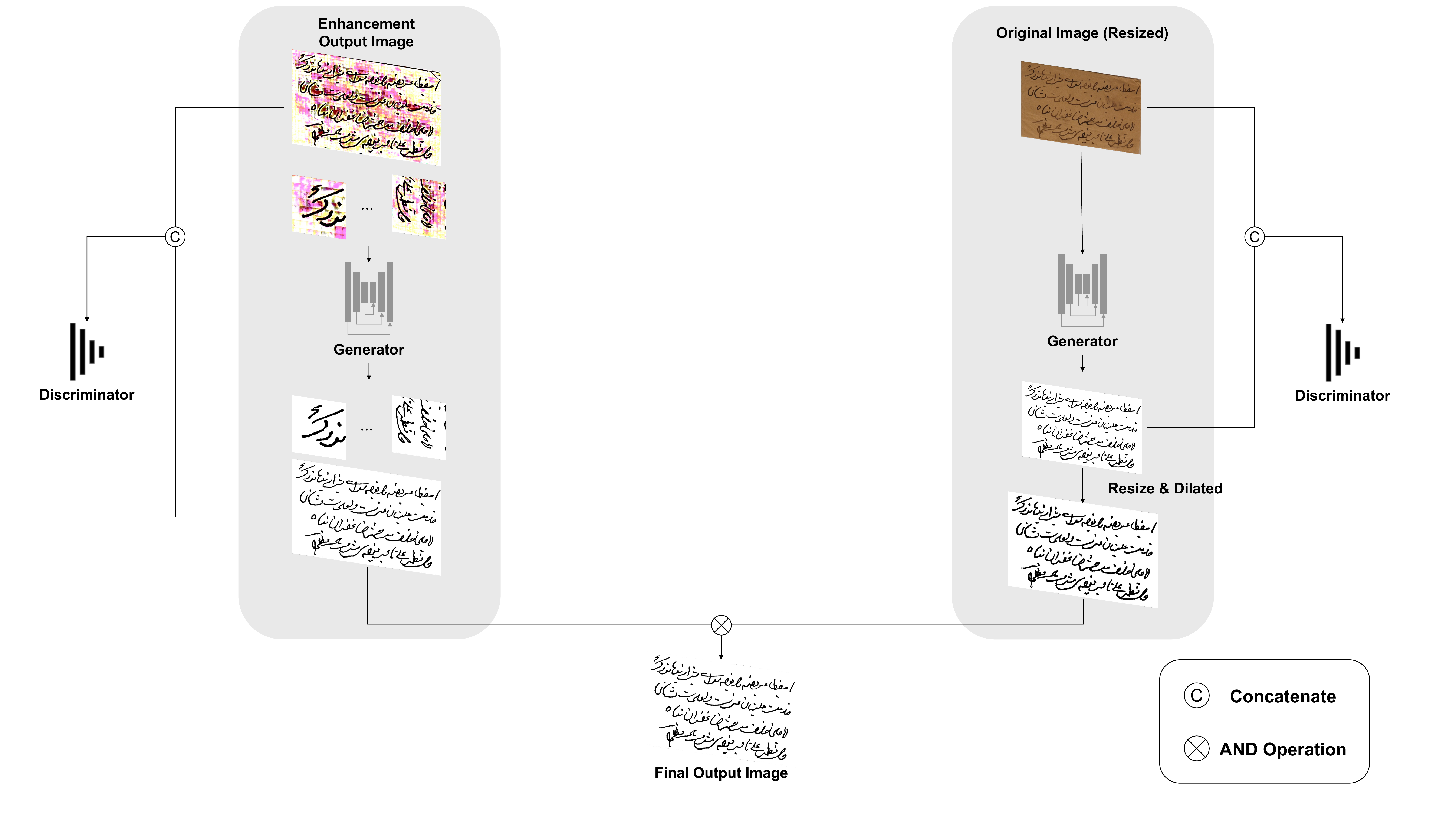}
  \caption{The structure of the proposed model for image binarization. The input image size for the left generator is $224\times224$ pixels, and for the right is $512\times512$ pixels.}
  \label{figure_stage_3}
  \vspace{-0.5cm}
\end{figure}

\subsection{Image Binarization}
\vspace{-0.2cm}
Finally, the proposed method employs a multi-scale adversarial network for generating images of both local and global binarization, enabling more accurate differentiation between the background and text. We conduct global binarization on the original input images to offset any potential loss of spatial contextual information in the images caused by local prediction. Since the input image for local prediction in this stage is an 8-bit image, and the image binarization stage employs a 24-bit three-channel image, we employ two independent discriminators in the image binarization stage, respectively. As depicted in Fig. \ref{figure_stage_3}, the input image for local prediction corresponds to the output of the image enhancement, while the input image size for global prediction is $512\times512$ pixels.

\subsection{Loss Function}
\vspace{-0.2cm}
In order to achieve a more stable convergence of the loss function, the proposed method utilizes the Wasserstein GAN \cite{gulrajani2017improved} target loss function. The report of Bartusiak \emph{et al.} \cite{bartusiak2019splicing} demonstrates that the binary cross-entropy (BCE) loss outperforms the L1 loss for binary classification tasks. Therefore, we utilize the BCE loss instead of the the L1 loss employed in Pix2Pix GAN \cite{isola2017image}. The Wasserstein GAN target loss function including the BCE loss is defined as follows:
\vspace{-0.5cm}

\begin{equation}
    \label{eq:theta_D}
        \mathbb{L}_D = -\mathbb{E}_{x,y}[D(y,x)] + \mathbb{E}_{x}[D(G(x), x)] + \alpha \mathbb{E}_{x, \hat{y}\sim P_{\hat{y}}}[(\Arrowvert \nabla_{\hat{y}} D(\hat{y}, x) \Arrowvert_2 -1 )^2]
\end{equation}

\vspace{-0.5cm}
\begin{equation}
    \label{eq:theta_G}
    \mathbb{L}_G = \mathbb{E}_{x}[D(G(x),x)] + \lambda \mathbb{E}_{G(x),y} [y\log G(x) + (1-y)\log (1-G(x))]	
\end{equation}
where the penalty coefficient is $\alpha$, and the uniform sampling along a straight line between the ground truth distribution $P_y$ and the point pairs of the generated data distribution is $P_{\hat{y}}$. $\lambda$ is used to control the relative importance of different loss terms. The parameter of the generator is $\theta_G$ and the parameter of the discriminator is $\theta_D$. In the discriminator, the generated image is distinguished from the ground truth image by the target loss function $\mathbb{L}_D$ in Eq. (\ref{eq:theta_D}). In the generator, the distance between the generated image and the ground truth image in each color channel is minimized by the target loss function $\mathbb{L}_G$ in Eq. (\ref{eq:theta_G}).

\begin{table}[ht]
\centering
\caption{Ablation study of the proposed model on benchmark datasets.}
\label{tab:ablation}
\setlength{\tabcolsep}{3.2mm}{
\begin{tabular}{cccccc}
\hline
\textbf{Methods} & \textbf{Dataset} & \textbf{FM↑} & \textbf{p-FM↑} & \textbf{PSNR↑} & \textbf{DRD↓} \\ \hline
Enhancement & DIBCO 2011 & 80.32 & 93.93 & 16.02 & 5.19 \\
Proposed &DIBCO 2011 & 94.08 & 97.08 & 20.51 & 1.75 \\ \hline
Enhancement & DIBCO 2013 & 86.19 & 97.36 & 17.91 & 3.81 \\
Proposed & DIBCO 2013 & 95.24 & 97.51 & 22.27 & 1.59 \\ \hline
Enhancement & H-DIBCO 2016 & 81.60 & 95.65 & 16.82 & 5.62 \\
Proposed & H-DIBCO 2016 & 91.46 & 96.32 & 19.66 & 2.94 \\ \hline
Enhancement & DIBCO 2017 & 78.76 & 93.30 & 15.15 & 5.84 \\
Proposed & DIBCO 2017 & 90.95 & 93.79 & 18.57 & 2.94 \\ \hline
\end{tabular}}
\vspace{-0.2cm}
\end{table}

\section{Experiments}
\label{experiments}
\subsection{Datasets}
\vspace{-0.2cm}
This work trains the model on several public datasets and compares the performance of the proposed method with other SOTA methods on benchmark datasets. Our training sets include Document Image Binarization Competition (DIBCO) 2009 \cite{gatos2009icdar}, Handwritten Document Image Binarization Competition (H-DIBCO) 2010 \cite{pratikakis2010h}, H-DIBCO 2012 \cite{pratikakis2012icfhr}, Persian Heritage Image Binarization Dataset (PHIBD) \cite{nafchi2013efficient}, Synchromedia Multispectral Ancient Document Images Dataset (SMADI) \cite{hedjam2013historical}, and Bickley Diary Dataset \cite{deng2010binarizationshop}. The test sets comprise DIBCO 2011 \cite{ratikakis2011icdar2011}, DIBCO 2013 \cite{pratikakis2013icdar}, H-DIBCO 2016 \cite{pratikakis2016icfhr2016}, and DIBCO 2017 \cite{pratikakis2017icdar2017}.
\vspace{-0.5cm}

\subsection{Evaluation Metric}
\label{metric}
\vspace{-0.2cm}
Four evaluation metrics are employed to evaluate the proposed method and conduct a quantitative comparison with other SOTA methods for document image binarization. The evaluation metrics utilized include F-measure (FM), Pseudo-F-measure (p-FM), Peak signal-to-noise ratio (PSNR), and Distance reciprocal distortion (DRD).
\vspace{-0.2cm}

\subsection{Experiment Setup}
\vspace{-0.2cm}
The backbone neural network of this work is EfficientNet-B6 \cite{tan2019efficientnet}. This paper utilizes a pre-trained model on the ImageNet dataset to reduce computational costs. During the image preprocessing stage, we divide the input images into $224\times224$ pixels patches, corresponding to the image size in the ImageNet dataset. The patches are sampled with scale factors of 0.75, 1, 1.25, and 1.5, and the images are rotated by 90°, 180°, and 270°. In total, the number of the training image patches are 336,702.

During the global binarization, we resize the original input image to $512\times512$ pixels and generate 1,890 training images by applying horizontal and vertical flips. The input images for the local binarization of the image binarization stage are obtained from the image enhancement stage, and both stages share the same training parameters. The image binarization stage is trained for 150 epochs, while the other stages are trained for 10 epochs each. This work utilizes the Adam optimizer with a learning rate of $2\times10^{-4}$. $\beta_1$ of the generator and $\beta_2$ of the discriminator are 0.5 and 0.999, respectively.
\vspace{-0.2cm}

\begin{figure}[ht]
  \centering
  \includegraphics[width=\linewidth]{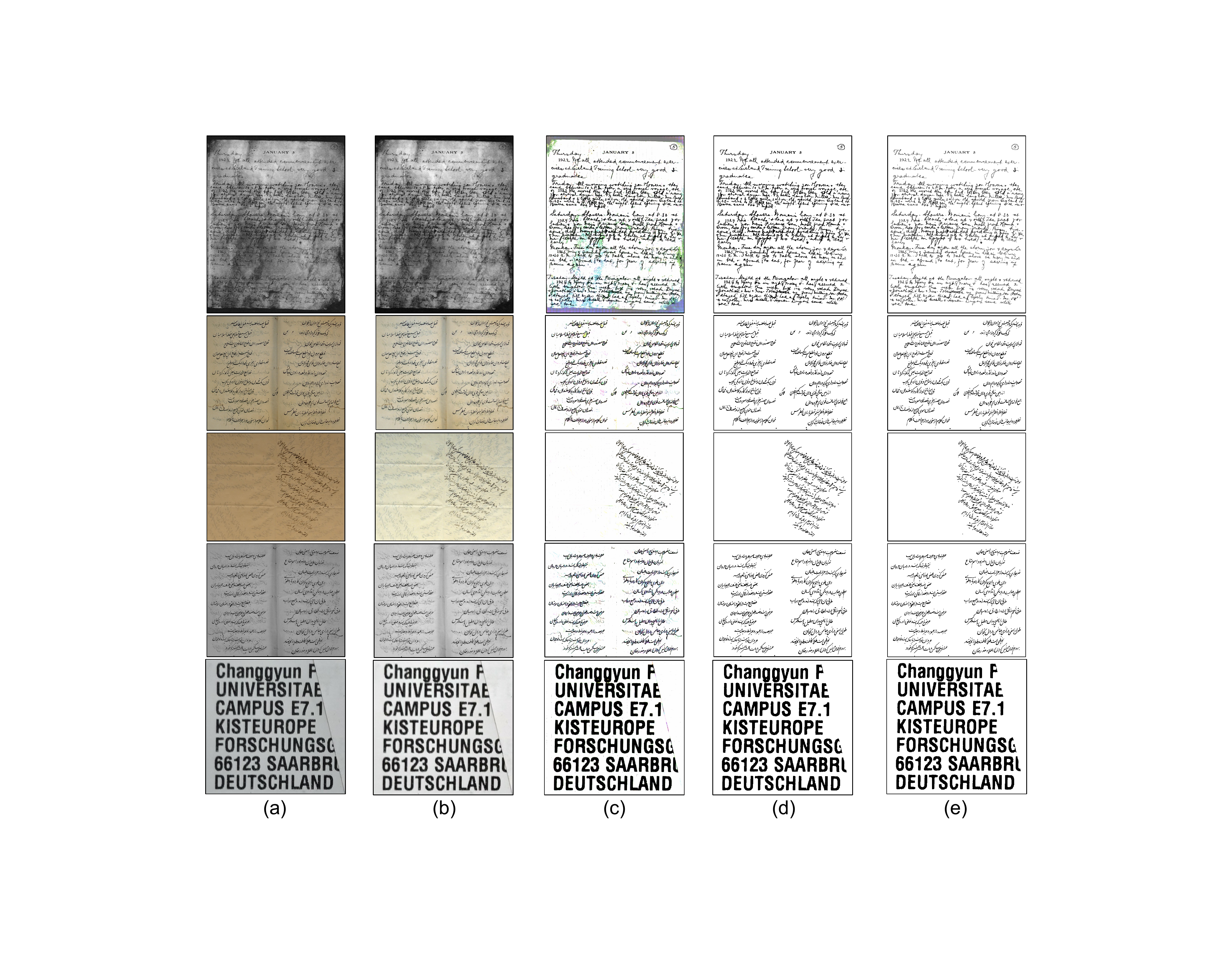}
  \caption{The output images of each stage of the proposed model: (a) the original input image, (b) the LL subband image after applying DWT and normalization, (c) the enhanced image using image enhancement method, (d) the binarization image using the method combining local and global features, (e) the ground truth image.}
  \label{figure_enhance}
  \vspace{-0.5cm}
\end{figure}

\subsection{Ablation Study}
\vspace{-0.2cm}
In this section, this work presents an ablation study conducted to assess the individual contributions of each stage of the proposed method. We evaluate the output of the image enhancement stage, as ``Enhancement'', and compare it with the final output, as ``Proposed''. The evaluation and comparison of the output results are performed on four DIBCO datasets. Table \ref{tab:ablation} demonstrates that the output result of ``Enhancement'' is worse than the final output in terms of FM, p-FM, PSNR, and DRD values.

To further demonstrate the advantages of each stage more intuitively, we choose five images from PHIBD \cite{nafchi2013efficient} and Bickley Diary Dataset \cite{deng2010binarizationshop} to show the step-by-step output results of image enhancement and image binarization using the proposed method. As shown in Fig. \ref{figure_enhance}, (b) represents the result of retaining the LL subband image after applying DWT and normalization (the result of the image preprocessing stage), showing that the original input image is performed noise reduction. (c) is the result of image enhancement using adversarial network, and it has removed the background color and highlighted the text color. (d) is the final output image obtained using the proposed method, and it can be seen that our final output is closer to the ground truth image (e).

\begin{table}[ht]
\centering
\vspace{-0.3cm}
\caption{Model performance comparision of different input images and ground truth images of the generator. Best and 2nd best performance are in {\color{red}red} and {\color{blue}blue} colors, respectively.}
\label{tab:comparison}
\subfigure[DIBCO 2011]{
\resizebox{1\columnwidth}{!}{
\begin{tabular}{ccccccc}
\hline
\textbf{Option} & \textbf{Input} & \textbf{GT} & \textbf{FM↑} & \textbf{p-FM↑} & \textbf{PSNR↑} & \textbf{DRD↓} \\ \hline
{1} & \textbackslash{} & \textbackslash{} & 86.68 & 89.61 & 19.27 & 4.01 \\ \hline
{2} & \textbackslash{} & DWT (LL) & 88.20 & 90.57 & 19.53 & 3.45 \\ \hline
{3} & \textbackslash{} & DWT (LL) + Norm & 87.70 & 90.24 & {\color{blue}19.65} & 3.45 \\ \hline
{4} & DWT (LL) & \textbackslash{} & 87.74 & 89.69 & 18.88 & 3.78 \\ \hline
{5} & DWT (LL) + Norm & \textbackslash{} & {\color{blue}89.33} & 91.94 & 19.49 & 3.37 \\ \hline
{6} & DWT (LL) & DWT (LL) & {\color{red}90.53} & {\color{red}92.82} & {\color{red}19.68} & {\color{red}3.11} \\ \hline
{7} & DWT (LL) + Norm &DWT (LL) + Norm & 89.06 & {\color{blue}92.25} & 19.59 & {\color{blue}3.31} \\ \hline
\end{tabular}}}
\subfigure[DIBCO 2013]{
\resizebox{1\columnwidth}{!}{
\begin{tabular}{ccccccc}
\hline
\textbf{Option} & \textbf{Input} & \textbf{GT} & \textbf{FM↑} & \textbf{p-FM↑} & \textbf{PSNR↑} & \textbf{DRD↓} \\ \hline
{1} & \textbackslash{} & \textbackslash{} & 92.94 & 94.70 & 21.57 & 2.74 \\ \hline
{2} & \textbackslash{} & DWT (LL) & 94.43 & 95.64 & 21.79  & 2.13 \\ \hline
{3} & \textbackslash{} & DWT (LL) + Norm & {\color{red}94.88} & {\color{red}96.19} & {\color{red}22.32} & {\color{red}1.95} \\ \hline
{4} & DWT (LL) & \textbackslash{}  & 93.23 & 94.43 & 20.80 & 2.67 \\ \hline
{5} & DWT (LL) + Norm & \textbackslash{} & 93.76 & 95.41 & 21.54 & 2.40  \\ \hline
{6} & DWT (LL) & DWT (LL) & 94.39 & 95.34 & 21.91 & 2.26 \\ \hline
{7} & DWT (LL) + Norm & DWT (LL) + Norm & {\color{blue}94.55} & {\color{blue}95.86} & {\color{blue}22.02} & {\color{blue}2.07} \\ \hline
\end{tabular}}}
\subfigure[H-DIBCO 2016]{
\resizebox{1\columnwidth}{!}{
\begin{tabular}{ccccccc}
\hline
\textbf{Option} & \textbf{Input} & \textbf{GT} & \textbf{FM↑} & \textbf{p-FM↑} & \textbf{PSNR↑} & \textbf{DRD↓} \\ \hline
{1} & \textbackslash{} & \textbackslash{} & 90.74 & 94.46 & 19.39 & 3.30 \\ \hline
{2} & \textbackslash{} & DWT (LL) & 91.76 & 95.74 & 19.67 & 2.93 \\ \hline
{3} & \textbackslash{} & DWT (LL) + Norm & 91.49 & {\color{red}96.46} & {\color{blue}19.68} & {\color{blue}2.92} \\ \hline
{4} & DWT (LL) & \textbackslash{} & {\color{blue}91.86} & 94.95 & 19.62 & 2.99 \\ \hline
{5} & DWT (LL) + Norm & \textbackslash{} & 91.28 & {\color{blue}96.03} & 19.47 & 3.04 \\ \hline
{6} & DWT (LL) & DWT (LL) & 91.68 & 95.90 & {\color{blue}19.68} & 2.93 \\ \hline
{7} & DWT (LL) + Norm & DWT (LL) + Norm & {\color{red}91.95} & 95.87 & {\color{red}19.75} & {\color{red}2.84} \\ \hline
\end{tabular}}}
\vspace{-0.5cm}
\end{table}

\begin{table}[ht]
\centering
\vspace{-0.3cm}
\caption{Quantitative comparison (FM/p-FM/PSNR/DRD) with other state-of-the-art models for document image binarization on benchmark datasets. Best and 2nd best performance are in {\color{red}red} and {\color{blue}blue} colors, respectively.}
\label{tab:result}
\subfigure[DIBCO 2011]{
\resizebox{0.48\columnwidth}{!}{
\begin{tabular}{ccccc}
\hline
\textbf{Methods} & \textbf{FM↑} & \textbf{p-FM↑} & \textbf{PSNR↑} & \textbf{DRD↓} \\ \hline
Otsu\cite{otsu1979threshold} & 82.10 & 85.96 & 15.72 & 8.95  \\ \hline
Sauvola\cite{sauvola2000adaptive} & 82.35 & 88.63 & 15.75 & 7.86  \\ \hline
He\cite{he2019deepotsu} & 91.92 & 95.82 & 19.49 & 2.37  \\ \hline
Vo\cite{vo2018binarization} & 92.58 & 94.67 & 19.16 & 2.38  \\ \hline
Zhao\cite{zhao2019document} & 92.62 & 95.38 & 19.58 & 2.55  \\ \hline
1st Place\cite{ratikakis2011icdar2011} & 88.74 & - & 17.97 & 5.36  \\ \hline
Yang\cite{yang2023gdb} & 93.44 & 95.82 & 20.10 & 2.25 \\ \hline
Suh\cite{suh2021cegan} & 93.44 & 96.18 & 19.97 & 1.93  \\ \hline
Tensmeyer\cite{tensmeyer2017document} & {\color{blue}93.60} & {\color{red}97.70} & {\color{blue}20.11} & {\color{blue}1.85}  \\ \hline
\textbf{Ours} & {\color{red}94.08} & {\color{blue}97.08} & {\color{red}20.51} & {\color{red}1.75} \\ \hline
\end{tabular}}}
\subfigure[DIBCO 2013]{
\resizebox{0.48\columnwidth}{!}{
\begin{tabular}{ccccc}
\hline
\textbf{Methods} & \textbf{FM↑} & \textbf{p-FM↑} & \textbf{PSNR↑} & \textbf{DRD↓} \\ \hline
Otsu\cite{otsu1979threshold} & 80.04 & 83.43 & 16.63 & 10.98 \\ \hline
Sauvola\cite{sauvola2000adaptive} & 82.73 & 88.37 & 16.98 & 7.34 \\ \hline
He\cite{he2019deepotsu} & 93.36 & 96.70 & 20.88 & 2.15 \\ \hline
Vo\cite{vo2018binarization} & 93.43 & 95.34 & 20.82 & 2.26 \\ \hline
Zhao\cite{zhao2019document} & 93.86 & 96.47 & 21.53 & 2.32 \\ \hline
1st Place\cite{pratikakis2013icdar}& 92.70 & 94.19 & 21.29 & 3.10 \\ \hline
Yang\cite{yang2023gdb} & {\color{blue}95.19} & 96.37 & {\color{red}22.58} & 1.78 \\ \hline
Suh\cite{suh2021cegan} & 94.75 & {\color{blue}97.36} & 21.78 & {\color{blue}1.73} \\ \hline
Tensmeyer\cite{tensmeyer2017document} & 93.10 & 96.80 & 20.70 & 2.20 \\ \hline
\textbf{Ours} & {\color{red}95.24} & {\color{red}97.51} & {\color{blue}22.27} & {\color{red}1.59} \\ \hline
\end{tabular}}}
\subfigure[H-DIBCO 2016]{
\resizebox{0.48\columnwidth}{!}{
\begin{tabular}{ccccc}
\hline
\textbf{Methods} & \textbf{FM↑} & \textbf{p-FM↑} & \textbf{PSNR↑} & \textbf{DRD↓} \\ \hline
Otsu\cite{otsu1979threshold} & 86.59 & 89.92 & 17.79 & 5.58 \\ \hline
Sauvola\cite{sauvola2000adaptive} & 84.27 & 89.10 & 17.15 & 6.09 \\ \hline
He\cite{he2019deepotsu} & {\color{blue}91.19} & {\color{blue}95.74} & {\color{blue}19.51} & {\color{blue}3.02} \\ \hline
Vo\cite{vo2018binarization} & 90.01 & 93.44 & 18.74 & 3.91 \\ \hline
Zhao\cite{zhao2019document} & 89.77 & 94.85 & 18.80 & 3.85 \\ \hline
1st Place\cite{pratikakis2016icfhr2016} & 88.72 & 91.84 & 18.45 & 3.86 \\ \hline
Guo\cite{guo2019nonlinear} & 88.51 & 90.46 & 18.42 & 4.13 \\ \hline
Bera\cite{bera2021non} & 90.43 & 91.66 & 18.94 & 3.51 \\ \hline
Yang\cite{yang2023gdb} & 90.41 & 94.70 & 19.00 & 3.34 \\ \hline
Suh\cite{suh2021cegan} & 91.11 & 95.22 & 19.34 & 3.25 \\ \hline
\textbf{Ours} & {\color{red}91.46} & {\color{red}96.32} & {\color{red}19.66} & {\color{red}2.94} \\ \hline
\end{tabular}}}
\subfigure[DIBCO 2017]{
\resizebox{0.48\columnwidth}{!}{
\begin{tabular}{ccccc}
\hline
\textbf{Methods} & \textbf{FM↑} & \textbf{p-FM↑} & \textbf{PSNR↑} & \textbf{DRD↓} \\ \hline
Otsu\cite{otsu1979threshold} & 77.73 & 77.89 & 13.85 & 15.54 \\ \hline
Sauvola\cite{sauvola2000adaptive} & 77.11 & 84.10 & 14.25 & 8.85 \\ \hline
Jia\cite{jia2018degraded} & 85.66 & 88.30 & 16.40 & 7.67 \\ \hline
Jemni\cite{jemni2022enhance} & 89.80 & 89.95 & 17.45 & 4.03 \\ \hline
Zhao\cite{zhao2019document} & 90.73 & 92.58 & 17.83 & 3.58 \\ \hline
1st Place\cite{pratikakis2017icdar2017} & {\color{blue}91.04} & 92.86 & 18.28 & 3.40 \\ \hline
Howe\cite{howe2013document} & 90.10 & 90.95 & {\color{blue}18.52} & 5.12 \\ \hline
Bera\cite{bera2021non} & 83.38 & 89.43 & 15.45 & 6.71 \\ \hline
Yang\cite{yang2023gdb} & {\color{red}91.33} & {\color{blue}93.84} & 18.34 & 3.24 \\ \hline
Suh\cite{suh2021cegan} & 90.95 & {\color{red}94.65} & 18.40 & {\color{red}2.93} \\ \hline
\textbf{Ours} & 90.95 & 93.79 & {\color{red}18.57} & {\color{blue}2.94} \\ \hline
\end{tabular}}}
\vspace{-0.6cm}
\end{table}

\begin{figure}[ht]
    \centering
    \includegraphics[width=\linewidth]{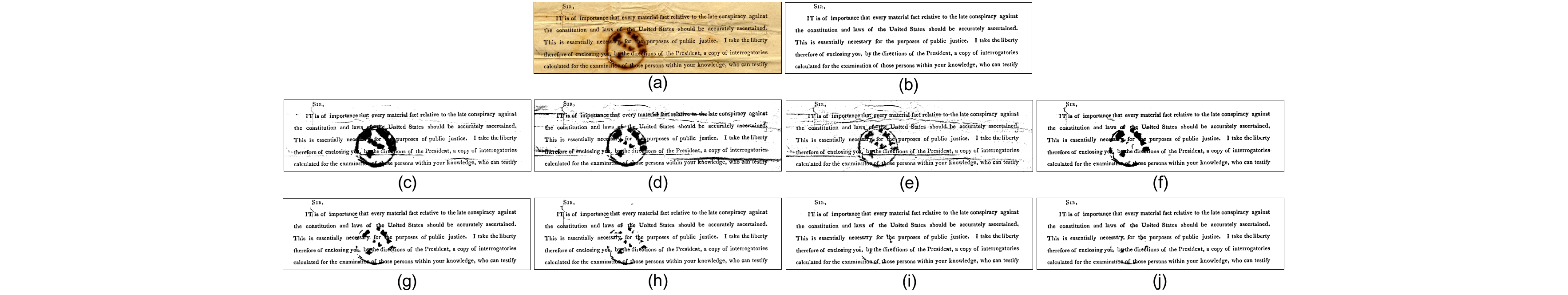}
    \vspace{-0.3cm}
    \caption{Examples of document image binarization for the input image PR16 of DIBCO 2013: (a) original input images, (b) the ground truth, (c) Otsu \cite{otsu1979threshold}, (d) Niblack \cite{niblack1985introduction}, (e) Sauvola \cite{sauvola2000adaptive}, (f) Vo \cite{vo2018binarization}, (g) He \cite{he2019deepotsu}, (h) Zhao \cite{zhao2019document}, (i) Suh \cite{suh2021cegan}, (j) Ours.}
    \label{figure_pr16}
\end{figure}

\begin{figure}[ht]
    \centering
    \includegraphics[width=\linewidth]{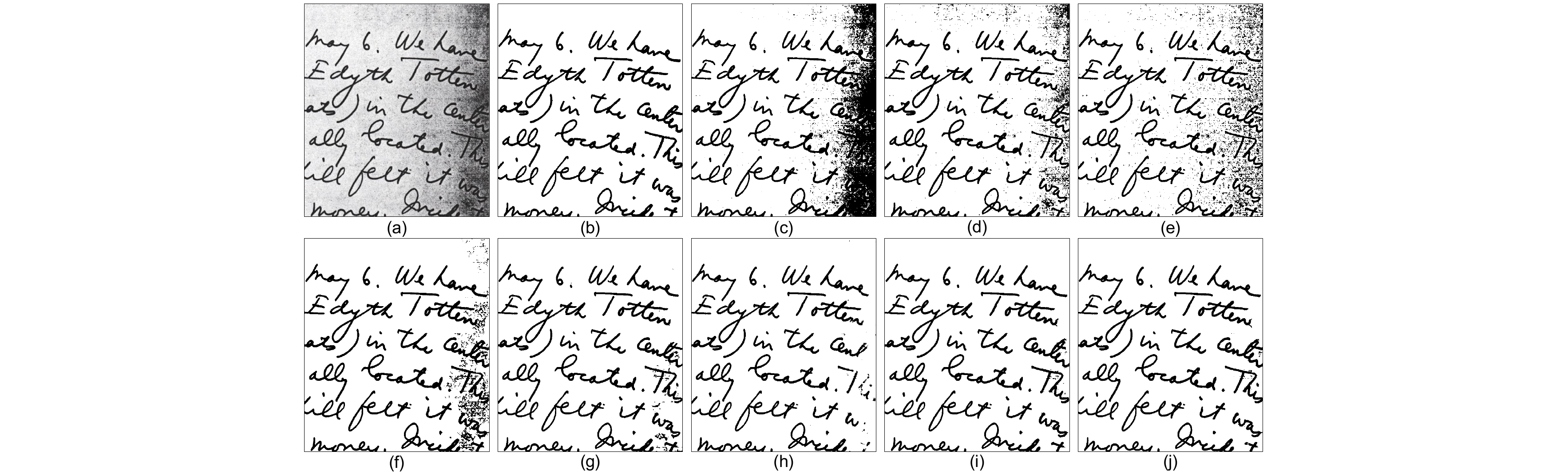}
    \vspace{-0.3cm}
    \caption{Examples of document image binarization for the input image HW5 of DIBCO 2013: (a) original input images, (b) the ground truth, (c) Otsu \cite{otsu1979threshold}, (d) Niblack \cite{niblack1985introduction}, (e) Sauvola \cite{sauvola2000adaptive}, (f) Vo \cite{vo2018binarization}, (g) He \cite{he2019deepotsu}, (h) Zhao \cite{zhao2019document}, (i) Suh \cite{suh2021cegan}, (j) Ours.}
    \label{figure_hw5}
    \vspace{-0.6cm}
\end{figure}

\subsection{Experimental Results}
\vspace{-0.2cm}
\label{experiment}
Despite mathematical theories supporting the effectiveness of applying DWT to images for storing contour information and reducing noise, we aim to comprehensively explain their impact on experimental results. To achieve this, we utilize UNet architecture \cite{ronneberger2015u} with EfficientNet-B5 \cite{tan2019efficientnet} as the baseline model to conduct comparison experiments, as presented in Table \ref{tab:comparison}. We formulate three options for the input images of the generator: direct input image, DWT to LL subband image, and DWT to LL subband image with normalization. Corresponding options are set up for the ground truth images. Notably, option 1: directly using the original input image as input, exhibits the worst performance on all four datasets. On DIBCO 2011 dataset, option 6: employing only DWT without normalization as the input image and corresponding to the ground truth image, demonstrate the best performance, achieving FM value of 91.95. The FM value of Option 3 reaches 94.88, achieving the top performance on DIBCO 2013 dataset by directly inputting the original image and utilizing the image processing output image as the corresponding ground truth image. Moreover, option 3 achieves the top two performance on DIBCO 2016 dataset. Based on this, we choose option 3 to employ UNet++ \cite{zhou2019unet++} with EfficientNet-B6 \cite{tan2019efficientnet} as the generator for network design.

Due to the lack of optical character recognition (OCR) result within dataset, both the proposed method and other SOTA methods are evaluated using the four evaluation metrics described in Section \ref{metric}. The evaluation results on the benchmark datasets are presented in Table \ref{tab:result}. Our proposed method demonstrates superior performance across all four evaluation metrics on DIBCO 2016 dataset. Additionally, on DIBCO 2011 and 2013 datasets, the proposed method achieves the top two performance in each evaluation metric. Despite slightly lower FM value of 90.05 compared to the highest value of 91.33, and p-FM value of 93.79 lower than the highest value of 94.65, the PSNR and DRD values maintain top two performance on DIBCO 2017 dataset. By combining the comparison results from these four datasets, it is demonstrated that the images produced by our proposed method exhibit greater similarity to the ground truth images, and better binarization performance.

To compare the difference between images generated by the proposed method and other methods, two images are selected as examples. Fig. \ref{figure_pr16} and Fig. \ref{figure_hw5} illustrate the results using different methods. Evidently, the proposed method preserves greater textual content while effectively eliminating shadows and noise compared to other methods.
\vspace{-0.3cm}

\section{Conclusion}
\vspace{-0.3cm}
\label{conclusion}
To perform image binarization on color degraded documents, this work splits the RGB three-channel input image into three single-channel images, and train the adversarial network on each single-channel image, respectively. Moreover, this work applies DWT on 224$\times$224 patches of single-channel image in the image preprocessing stage to improve the model performance. We name the proposed generative adversarial network as CCDWT-GAN, which achieves SOTA performance on multiple benchmark datasets. 

\subsubsection{Acknowledgment}
\vspace{-0.3cm}
This work is supported by National Science and Technology Council of Taiwan, under Grant Number: NSTC 112-2221-E-032-037-MY2.

\vspace{-0.3cm}
\bibliographystyle{splncs04}
\bibliography{mybibliography}

\end{document}